# Power Plant Performance Modeling with Concept Drift

Rui Xu
Machine Learning Laboratory
GE Global Research
Niskayuna, NY, USA
xur@ge.com

Yunwen Xu
Machine Learning Laboratory
GE Global Research
San Ramon, CA, USA
yunwen.xu@ge.com

WeiZhong Yan
Machine Learning Laboratory
GE Global Research
Niskayuna, NY, USA
yan@ge.com

*Abstract*—**Power plant is a complex and nonstationary system for which the traditional machine learning modeling approaches fall short of expectations. The ensemble-based online learning methods provide an effective way to continuously learn from the dynamic environment and autonomously update models to respond to environmental changes. This paper proposes such an online ensemble regression approach to model power plant performance, which is critically important for operation optimization. The experimental results on both simulated and real data show that the proposed method can achieve performance with less than 1% mean average percentage error, which meets the general expectations in field operations.**

*Keywords—performance modeling; ensemble learning; online learning; learning in nonstationary environments; concept drift*

## I. INTRODUCTION

In today's competitive business environment, power plant owners are constantly striving to reduce their operation and maintenance costs, thus increasing their profits. To enable plant owners to operate their plants more efficiently, it is important to develop advanced digital solutions (software and tools) that can provide decision support for the plant operation optimization. For example, Digital Power Plant, a part of the GE's vision for the digitization of industrial assets, is one of such technologies recently developed in GE. Digital Power Plant involves building a collection of digital models (both physics-based and data-drive), or "Digital Twins" as we call it at GE, which are used to model the present state of every asset in a power plant. This transformational technology enables utilities to monitor and manage every aspect of the power generation ecosystem to generate electricity as cleanly, efficiently, and securely.

Power plant is an inherently dynamic system due to the physics driven degradation, different operation and control settings, and various maintenance actions. For example, the efficiency of an asset or equipment degrades gradually because of part wearing due to part aging, rubbing between stationary and rotating parts, and so on. External factors, such as dust, dirt, humidity, and temperature can also affect the characteristics of these assets or equipment. The change of operation condition may cause unseen scenarios in observed data. For example, for a combined cycle power plant, the on-off switch of a duct burner will lead to the relationship change between the power output and the corresponding input variables. The maintenance actions, particularly online actions, will usually cause abrupt changes to the system behavior. A typical example is water wash of compressor, which could significantly increase its efficiency and lead to higher power output under similar environments.

Learning in nonstationary environments, also known as concept drift learning or learning in dynamics in the literature, has attracted lots of efforts for the past decades, particularly in the context of classification in the communities of machine learning and computational intelligence [1-5]. It also has close relations with many other research areas, such as transfer learning [6], Kalman filter [7], multitask learning [8], stream and time series mining [9], and so on. Basically, concept drift can be distinguished to two types – real drift, which refers to the change of the posterior probability, and virtual drift, which refers to the change of prior probability without affecting the posterior probability [1-3]. The physical system degradation and operation condition change are real drifts. Insufficient data representation for initial modeling belongs to virtual drift. Concept drift can also be classified into three types of patterns based on the change rate over time [1-3]. Sudden drift indicates the drift happens abruptly from one concept to another (e.g., water wash of power gas turbine can increase the compressor efficiency - a hidden variable, which leads to the significant increase of power output). In contrast to sudden drift, gradual drift takes a longer period for concept evolving (e.g., the wear of parts leads to the degradation of a physical system). The drift can also be recurring with the reappearance of the previous concept.

Generally, the adaptation algorithms for concept drift belong to two primary families – active approaches and passive approaches, based on whether an explicit detection of change in the data is required or not [1]. For the active approaches, the adaptation mechanism is only triggered after the change is detected. In contrast, passive approaches continuously learn over time, assuming that the change can happen at any time with any change pattern or rate.

Under the framework of active approaches, the drift detection algorithms monitor either the performance metrics or the characteristics of data distribution, and notify the adaptation mechanism to react to detected changes. Commonly used detection technologies include sequential hypothesis test, change detection test, and hypothesis tests [1, 10-11]. The major challenge to the adaptation mechanisms is to select the most relevant information to update the model. A simple strategy is to apply a sliding window, and only data points within the current window are used to retrain the model. The window size can be fixed in advance or adjusted adaptively [12-14]. Instance weighting is another approach to address this problem, which assigns weights to data points based on their age or relative importance to the model performance [15]. Instance weighting requires the storage of all previous data, which is infeasible for many applications with big data. An alternative approach is to apply data sampling to maintain a data reservoir that provides training data to update the model [16].

Passive approaches perform continuous update of the model upon the arrival of new data points. It is closely related to the research topics of continuous learning and online learning. The continuously evolving learner can be either a single model or an ensemble of models. The latter has become more popular in recent decade thanks to its inherent advantages to single model. Particularly, ensemble-based learning provides a very flexible structure to add and remove models from the ensemble, thus providing an effective balance in learning between new and old knowledge. There are lots of ensemble-based passive algorithms proposed in the literature [17-24], and they vary from the following aspects,

- Voting strategy – Weighted voting is a common choice for many algorithms [17, 18, 20-24], but some authors argue the average voting might be more appropriate for nonstationary environment learning [19]. Reference [20] introduces three dynamic techniques for model integration.

- Voting weights – If weighted voting is used, the weighs are usually determined based on the model performance. For example, in [17], the weight for each learner is calculated as the difference of mean square errors between a random model and the learner. The algorithm DWM (Dynamic Weighted Majority) penalized wrong prediction of the learner by decreasing the weight with a pre-determined factor [18]. The weight for each leaner is calculated as the log-normalized reciprocals of the weighted errors in the algorithm Learn++.NSE [21].

- New model – When and how to add a new model to the ensemble is critically important to the effective and fast adaptation to the environment changes. References [17] and [22] build a new model for every new chunk of data. More commonly, a new model is added if the ensemble performance on the current data point(s) is wrong or below expectation [18, 23, 24]. The training data usually are the most recent samples [20, 23].

- Ensemble pruning – In practice, the ensemble size is usually bounded due to the limitation of resources. A simple pruning strategy is to remove the worst performance model whenever the upper bound of the ensemble is reached [22, 23]. The effective ensemble size can also be dynamically determined by approaches, such as instance based pruning [17] and ordered aggregation [24]. The algorithm DWM removes a model from the ensemble if its weight is below a threshold [19].

The most recent advances on learning in streaming data with imbalanced classes under nonstationary environments are reported in [25] and [26]. An ensemble-based online learning algorithm is proposed to address the problem of class evolution, i.e., the emergence and disappearance of classes with the streaming data [27].

In this paper, we focus on the application of ensemble-based passive approach for the power plant performance modeling. The major difference between our work and some previous efforts for power plant modeling [28] is that we consider modeling in dynamic environments. The algorithm we applied is mostly based on the DOER algorithm (Dynamic and On-line Ensemble Regression) proposed in [23], considering its overall better performance on multiple synthetic and real (industry applications) data sets when compared to several state-of-the-art algorithms. For the same reason, we will not provide comparison to other approaches in this paper. We make some modifications to DOER based on the specific requirements of this application. For example, we add a long-term memory, based on reservoir sampling, to store previous knowledge, and select the most similar data points from the long-term memory and the current data as the training set for a new model. Such a change is effective to make the algorithm adapt to abrupt change in a faster way, as for a sudden change, the data points before the change point are no longer the representatives of the real information. For power plant modeling, it is a common phenomenon that the compressor or turbine efficiency improves because of either online or offline water wash. Such an improvement will lead to a sudden increase of power output in general. We also extend DOER for problems with multiple outputs. Like DOER, we also use the online sequential extreme learning machines (OS-ELM) [29] as the base model in the ensemble, which is an online realization of ELM. OS-ELM enjoys the advantage of very fast training and easiness for implementation.

The remainder of this paper is organized as follows. Section II describes the ensemble-based online regression approach for power plant performance modeling. In Section III, we discuss how data are prepared and present the empirical results on the simulated and real plant data. Section IV concludes this paper.

II. METHODS

*A. A Brief on ELM and OS-ELM*

Extreme learning machine (ELM) is a special type of feed-forward neural networks introduced by Huang, et al. [30].

Unlike in traditional feed-forward neural networks where training the network involves finding all connection weights and bias, in ELM, connections between input and hidden neurons are randomly generated and fixed, that is, they do not need to be trained. Thus, training an ELM becomes finding connections between hidden and output neurons only, which is simply a linear least squares problem whose solution can be directly generated by the generalized inverse of the hidden layer output matrix [30]. Because of such special design of the network, ELM training becomes very fast. Numerous empirical studies and recently some analytical studies as well have shown that ELM has better generalization performance than other machine learning algorithms including SVMs, and is efficient and effective for both classification and regression [30,31].

Consider a set of $N$ training samples, $\{(\boldsymbol{x}_i, \boldsymbol{y}_i)\}_{i=1}^N$, $\mathbf{x}_i \in \Re^d$, $\mathbf{y}_i \in \Re^r$. Assume the number of hidden neurons is $L$. Then the output function of ELM for generalized single layer feed-forward neural networks is,

$$f(\boldsymbol{x}) = \sum_{i=1}^{L} \boldsymbol{\beta}_i h_i(\boldsymbol{x}) = \boldsymbol{H}(\boldsymbol{x})\boldsymbol{\beta} \quad (1)$$

where $\boldsymbol{h}_i(\boldsymbol{x}) = G(w_i, b_i, \boldsymbol{x})$, $w_i \in \Re^N, b_i \in \Re^r$, is the output of $i^{th}$ hidden neuron with respect to the input $\boldsymbol{x}$; $G(w, b, \boldsymbol{x})$ is a nonlinear piecewise continuous function satisfying ELM universal approximation capability theorems [31]; $\boldsymbol{\beta}_i$ is the output weight matrix between $i^{th}$ hidden neuron to the $k \geq 1$ output nodes. $\boldsymbol{H}(\boldsymbol{x}) = [\boldsymbol{h}_1(\boldsymbol{x}), ..., \boldsymbol{h}_L(\boldsymbol{x})]$ is a random feature map mapping the data from $d$-dimensional input space to the $L$-dimension random feature space (ELM feature space).

For batch ELM where all samples are available for training, the output weight vector can be estimated as the least-squares solution of $\boldsymbol{H}\boldsymbol{\beta} = \boldsymbol{Y}$, that is, $\widehat{\boldsymbol{\beta}} = \boldsymbol{H}^\dagger \boldsymbol{Y}$, where $H^\dagger$ is the Moore-Penrose generalized inverse of the hidden layer output matrix (see [30] for details), which can be calculated through the orthogonal projection method:

$$\boldsymbol{H}^\dagger = (\boldsymbol{H}^T \boldsymbol{H})^{-1} \boldsymbol{H}^T \quad (2)$$

Online sequential ELM (OS-ELM), proposed by Liang, et al. [29], is a variant of classical ELM, which has the capability of learning data one-by-one or chunk-by-chunk with a fixed or varying chunk size. As described in details in [29], OS-ELM involves two learning phases, *initial training* and *sequential learning*.

**Phase I - Initial training**: choose a small chunk of initial training samples, $\{(\boldsymbol{x}_i, \boldsymbol{y}_i)\}_{i=1}^{M_0}$, where $M_0 \geq L$, from the given $M$ training samples; and calculate the initial output weight matrix, $\boldsymbol{\beta}^0$, using the batch ELM formula described above.

**Phase II - Sequential learning**: for $(M_0 + k + 1)^{th}$ training sample, perform the following two steps.

1) Calculate the partial hidden layer output matrix:

$$\boldsymbol{H}_{k+1} = [h_1(x_{M_0+k+1}), ..., h_L(x_{M_0+k+1})], \quad (3)$$

and set

$$\boldsymbol{t}_{k+1} = \boldsymbol{y}^T_{(M_0+k+1)}. \quad (4)$$

2) Calculate the output weight matrix:

$$\boldsymbol{\beta}^{k+1} = \boldsymbol{\beta}^k + \boldsymbol{R}_{k+1}\boldsymbol{H}_{k+1}(\boldsymbol{t}^T_{k+1} - \boldsymbol{H}^T_{k+1}\boldsymbol{\beta}^k), \quad (5)$$

where,

$$\boldsymbol{R}_{k+1} = \boldsymbol{R}_k - \frac{R_k H_{k+1} H^T_{k+1} R_k}{1 + H^T_{k+1} R_k H_{k+1}} \quad (6)$$

for $k = 0, 1, 2, ..., M - M_0 + 1$.

### B. Online Ensemble Learning

The algorithm we applied for the power plant performance modeling consists of two major phases (see Fig. 1): the initialization phase and the online learning phase, which includes two major steps, i.e., model performance evaluation and model set update.

**Phase I: Initial Training**.

During the initialization phase, the algorithm creates the first model, $m_1$, based on the given initial training data $\mathbf{D}_{init} = \{(\mathbf{x}_t, \mathbf{y}_t) \mid t = 1, ..., T_I, \mathbf{x}_t \in \Re^d, \mathbf{y}_t \in \Re^r\}$, where $d \geq 1$ and $r \geq 1$ are the dimensions for input and output variables, respectively. The number of hidden nodes of the OS-ELM is determined based on k-fold cross validation on $\mathbf{D}_{init}$. The algorithm then maintains two data windows with fixed size $ws$. The first data window is called short term memory $\mathbf{D}_S$, which contains the most recent $ws$ data points from the stream. The other data window is known as long term memory $\mathbf{D}_L$, which collects data points from the stream based on reservoir sampling [16]. Specifically, this sampling strategy initially takes the first $ws$ data points to the reservoir. Subsequently, the $t$ data point is added to the reservoir with the probability $ws / t$. A randomly selected point is then removed from the reservoir. For the data point that leads to the creation of a new model, the probability for keeping it in the reservoir is 1. By maintaining both long and short term memories, we expect that the algorithm can take advantage of both the previous and most recent knowledge.

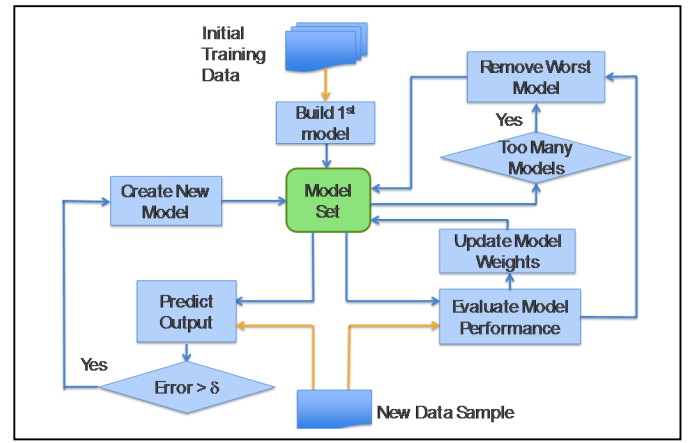

Fig. 1. A flowchart of the online, dynamic, and ELM-based ensemble regression algorithm.

Each model is associated with a variable, named *Life*, which counts the total number of online evaluations the model has seen so far. It is easy to see that *Life* should be initialized as 0 for each new model. The mean square error (MSE) of the model on the data points that it is evaluated on (with upper threshold ≤ *ws*) is denoted as a variable, *mse*, which is also initially set as 0. The voting strategy of the ensemble is weighted voting, and the weight of the first model is 1.

**Phase II: Online Learning**.

In the online learning phase, the ensemble generates the prediction $\widehat{y_t}$ for a new input point $\mathbf{x}_t$, based on weighted voting from all of its components,

$$\widehat{y_t} = \sum_{i=1}^{M} w_i \mathbf{o}_i / \sum_{i=1}^{M} w_i \qquad (7)$$

where *M* is the total number of models in the ensemble, $w_i$ is the weight of the model $m_i$, and $\mathbf{o}_i$ is the output from the model $m_i$.

Correspondingly, the prediction error of model $m_i$ on the new data point is obtained as,

$$e_i^t = \sum_{j=1}^{r} (y_t^j - o_t^j)^2 \qquad (8)$$

For each model $m_i$, its weight is adjusted based on $mse_i$. With the calculated squared error $e_i^t$, the variable $mse_i$ is calculated as,

$$mse_i^t = \begin{cases} 0 & \text{if } life_i = 0 \\ \frac{life_i - 1}{life_i} \times mse_i^{t-1} + \frac{e_i^t}{life_i} & \text{if } 1 \leq life_i \leq ws \\ mse_i^{t-1} + \frac{e_i^t}{ws} - \frac{e_i^{t-ws}}{ws} & \text{if } life_i > ws \end{cases} \qquad (9)$$

Accordingly, the weight $w_i$ for the model $m_i$ is updated as,

$$w_i = e^{-\left(\frac{mse_i^t - median(\Psi^t)}{median(\Psi^t)}\right)} \qquad (10)$$

where $\Psi^t = (mse_1^t, \ldots, mse_m^t)$ is the set of the MSEs of all models in the ensemble and $median(\Psi^t)$ takes the median of MSEs of all models. As shown in (10), the impact of a model on the ensemble output decreases exponentially with its MSE larger than the median. On the other hand, models with smaller MSEs than the median will contribute more to the final ensemble output.

Following the weight updates, the models in the ensemble are all retrained by using the new point $(\mathbf{x}_t, \mathbf{y}_t)$, based on the updating rules of OS-ELM.

To determine whether a new model is needed to be added to the ensemble, the algorithm evaluates the absolute percentage error of the ensemble on the new point $(\mathbf{x}_t, \mathbf{y}_t)$,

$$APE_j = abs\left(\frac{\widehat{y_j} - y_j}{y_j}\right) \times 100, \ j = 1, \ldots, r \qquad (11)$$

If $APE_j$ ($j = 1, \ldots, r$) is greater than a threshold $\delta_j$, a new model is going to be created. In other words, if the ensemble fails to achieve pre-determined accuracy on either of the output, a new model will be added to the ensemble. Note that the thresholds could be different for different outputs based on the specific requirements. Like the process for building the first model, the variables *Life* and *mse* for the new model are set to 0, and the weight assigned to the model is 1.

The training data for the new model are selected from the long term and short term memories, i.e., $\mathbf{D}_L$ and $\mathbf{D}_S$, based on how similar between the points in these two sets and the new data point $(\mathbf{x}_t, \mathbf{y}_t)$. To calculate such distances, both input and output variables are considered, which leads to an extension vector $\mathbf{z} = (\mathbf{x}, \mathbf{y}) = (x_1, \ldots, x_d, y_1, \ldots, y_r)$. Given the candidate set combined from $\mathbf{D}_L$ and $\mathbf{D}_S$, i.e., $\mathbf{D}_C = (\mathbf{z}_1, \ldots, \mathbf{z}_{2 \times ws})$, and the current data point $\mathbf{z}_t = (\mathbf{x}_t, \mathbf{y}_t)$, the distance between $\mathbf{z}_t$ and $\mathbf{z}_j \in \mathbf{D}_C$ is calculated as,

$$dis(\mathbf{z}_t, \mathbf{z}_j) = \sum_{k=1}^{d} W_k (x_t^k - x_j^k)^2 + \sum_{l=1}^{r} W_{d+l} (y_t^l - y_j^l)^2 \qquad (12)$$

where $\mathbf{W} = (W_1, \ldots, W_{d+r})$ are the weights for the input and output variables. For this study, we assign (5 times) larger weights to the output variables than input variables to emphasize the impact of hidden factors, such as operation conditions and component efficiency.

We then define a threshold τ as the mean of all these distances minus the standard deviation. All candidate points from $\mathbf{D}_C$ with their distances to the current data point less than τ are included in the training set. If the total number of points in the training set is too small, e.g., less than *ws*, we will add more candidate points to the training set based on the order of their distances to the current data point till the training set has *ws* data points.

For this algorithm, the maximum number of models in the ensemble is fixed. Therefore, if the number of models is above a threshold *ES* because of the addition of a new model, the worst performance model, in terms of the variable *mse*, will be removed from the ensemble.

After all the updates discussed above are done, the weights of the models are normalized.

III. EXPERIMENT RESULTS

*A. Data Set*

The data sets we used in this paper include both the simulated data and the real data. All the data sets include 9 input variables, known as compressor inlet temperature, compressor inlet humidity, ambient pressure, inlet pressure drop, exhaust pressure drop, inlet guide vane angle, fuel temperature, compressor flow, and controller calculated firing temperature. The output variables are the gross power output and net heat rate.

First, to investigate the algorithm performance on drift with different patterns and rates, we generated simulation data by adjusting the compressor efficiency, which is a hidden variable to the model. For example, Fig. 2 illustrates a simulated data set for water wash of engine and gradual wear-out of machine parts within 1-year range. As shown in the top plot of Fig. 2, the compressor efficiency first linearly decreases from 1 to 0.9, and then jumps to 1.1 at change point 40,000, which corresponds to the water wash of the engine. The compressor efficiency remains stable at 1.1 for 10,000 points, and decreases again. The compressor efficiency, together with the 9 input variables, which are obtained from the real plant, are then provided as the inputs to the GE Power simulation tool, known as GTP (Gas Turbine Performance). GTP generates the outputs of power output and heat rate for further analysis. As shown in the bottom plot of Fig. 2, it is clear to see the impact of the change of the compressor on the gross power output from GTP. Particularly, at the change point 40,000, the power output increases significantly because of the significant improvement of the compressor efficiency. There are also some noise or outliers with the data (e.g., data points with power output = 0), which are removed from further analysis.

To have some statistical favor, we also generate 500 simulated data series, each of which contains 2,000 data points that are a chunk of the data in Fig. 2. The generated sequences basically belong to 2 types of changes – sudden and gradual change (265 series with sudden change and 235 series with gradual change). For sudden change, the compressor efficiency starts at 1.0 and then gradually decreases to 0.9. It jumps to 1.1 at the change point, and decreases to 0.9, where it jumps again to 1.1. the efficiency will stay at 1.1 for a while and then gradually drops to 0.95. For gradual change, the compressor efficiency still starts at 1.0 and then gradually decreases to and stay at 0.9. The change point, change range, and stable range are randomly selected for each sequence.

For the real data set we evaluated, we directly use the base load gross power and the base load gross LHV heat rate from the plant. The date ranges we take for each sequence is from January 1, 2015 to May, 31, 2016. The data points are sampled every 5 minutes, and any record with missing values is removed.

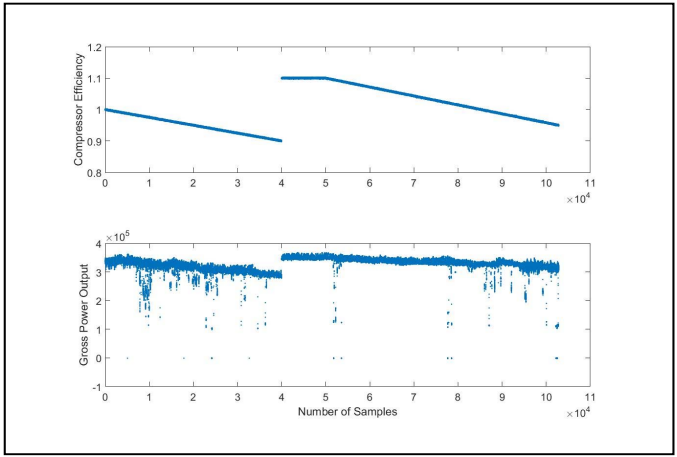

Fig. 2. Example of simulated data (compressor efficiency and gross power output.

## B. Results

To investigate the sensitivity of the algorithms to the parameters, particularly, the window size $ws$ and the threshold $\delta$ for adding a new model, we set $ws$ in the range of {100, 500, 1000, 1500, 2000, 3000, 4000, 5000}, and let $\delta$ vary from 0.01 to 0.1 with a step size of 0.01. Other parameters are fixed. The data set illustrated in Fig. 2 is used for this analysis after outliers are removed. As shown in Fig. 3, in general, the performance of the algorithm, measured in terms of mean absolute percentage error (MAPE), is better for smaller $\delta$. In other words, the threshold $\delta$ needs to be set to some small value to adapt fast to the changes. It also can be seen from the figure that the algorithm is not very sensitive to the window size $ws$ when $\delta$ is small. As $\delta$ becomes larger, either very small or large window lead to worse performance. In general, a window size of 500 or 1,000 is a good choice for good performance.

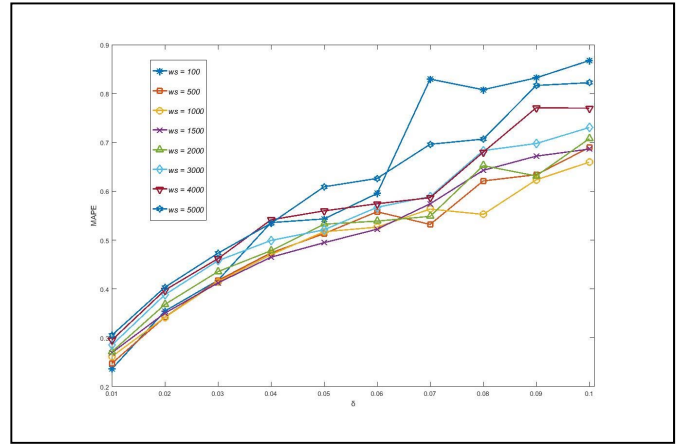

Fig. 3. Effect of the parameter $ws$ and $\delta$ on the algorithm performance.

Fig. 4 shows the influence of the maximum number of models, $ES$, on the algorithm performance for both the simulated data (top) (Fig. 2) and the real data (bottom). We run $ES$ in the range of 2 to 16, with the MAPE for each value obrained as the mean from 10 runs on the data set. $ws$ and $\delta$ are set at 1000 and 0.04, respectively. In general, there is no significnat performance change across the entire range investigated for $ES$. For the simulated data, the increase of $ES$ does not bring improvement to the performance, but for the real data, the performance becomes slightly better when $ES$ ranges from 6 to 12. The selection of $ES$ is problem dependent, however, values ranging in [6, 12] is a good start to make sure there are enough models in the ensemble while reducing computational burben or avoiding overcomplexities.

The performance of the proposed aglorithm on the 500 data series is depicted in Fig. 5. We select ELM without retraining and OS-ELM as the benchmark for comparsion, as they are good representatives of major practice for data-driven modeling in industry. The performance from the original DOER algorithm is also included. As we are more interested in how these algorithms repond when concept drift happens, for each series, we only calculate the MAPE for a subset of the series that starts from the 100 points before the change appears and lasts for the entire change range. We run each algorithms 5 times on each series, and the box-and-whisker plot is drawn based on the mean performance on the series. Particulary, the top plot in Fig. 5 shows the performance on the series with sudden change, while the bottom plot is for the series with gradual change. For both types of changes, it is obvious that the ELM without retraing and OS-ELM do not work well, with mean and standard deviation as 5.201±1.539 (sudden change) and 8.896±0.879 (gradual change), and 5.148±1.244 (sudden change) and 4.526±1.785 (gradual change), respectively. The MAPEs for the DOER are 2.219±1.790 (sudden change) and 1.370±1.420 (gradual change). In comparison, the MAPEs for the modified DOER are 2.116±1.681 (sudden change) and 1.546±1.506 (gradual change), which are slightly better for series with sudden changes, but deteriorate slightly for gradual change cases. The inclusion of LTM increase the algorithm's capability to faster adapt to sudden changes due to operation condition change or maintainance action, however, how to more effectively select samples for a new model training is still a problem needs further investigation. It is worthwhile to mention that the means and standard deviations of the proposed algorithm on the entire non-training series are 0.813±0.109 (sudden change) and 0.474±0.031 (gradual change), which meet 1% expectation in practice.

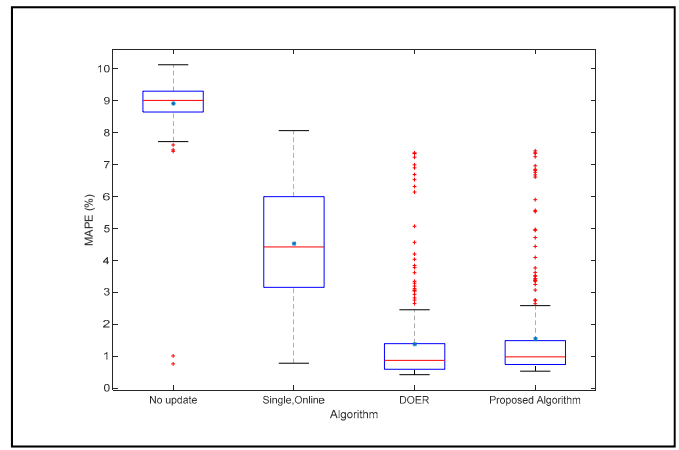

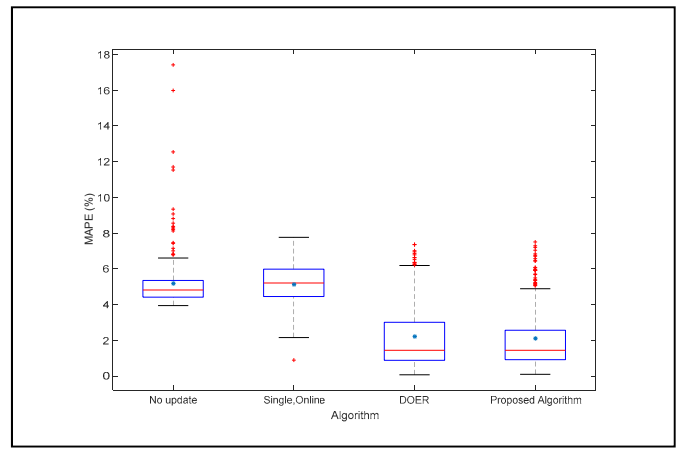

Fig. 5. Box-and-whisker plots that summarize the performance (MAPE) of 4 different approaches on the simulated data seires. The top is for gradual changes and the bottom is for sudden changes. The blue star ('*') represents the mean.

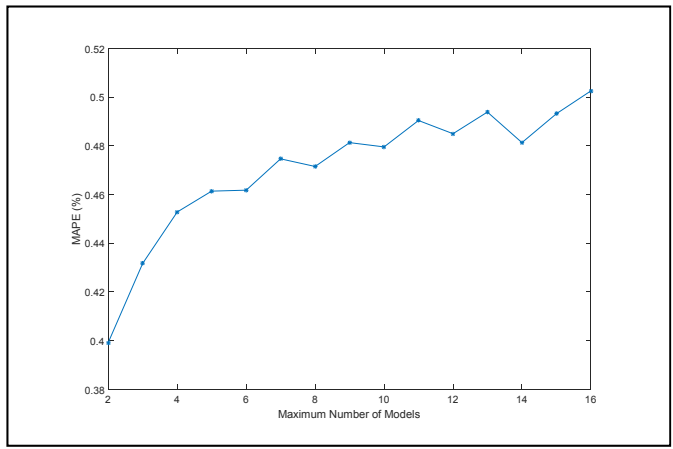

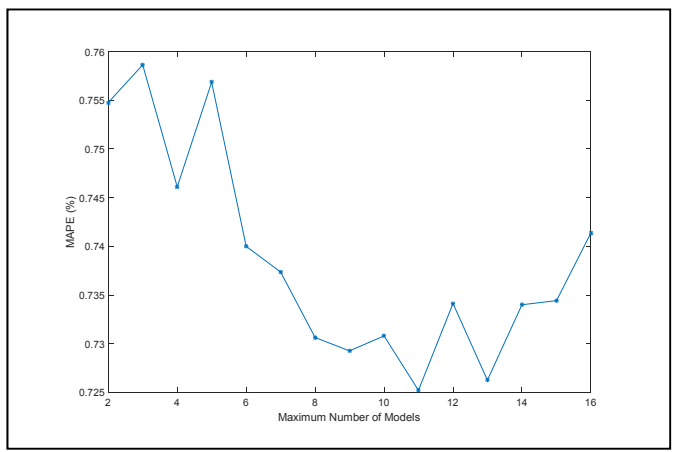

Fig. 4. Effect of the maximum number of models in the ensemble on the algorithm performance (top, simulated data; bottom, real data).

Similarly, we compare the performance of DOER and the proposed algorithm on the real data set when water wash happened, as maintenance action (either online or offline) is an important factor leading to concept drift. Specifically, Fig. 6 shows the box-and-whisker plots for both the power output and the heat rate. The means and standard deviations of MAPEs for the proposed algorithm on power output and heat rate are 1.114±0.067 and 0.615±0.034, respectively. In comparison, the DOER achieves 1.278±0.024 and 0.774±0.018 on these two outputs.

Fig. 7 illustrates an example on how the proposed algorithm respond to a maintenance action (water wash) at the $201^{st}$ sample point. The algorithm processed 13 more data points after the change happened to reach an error less than 1% (0.99%) for power output, and 4 samples to regain an error less than 1% (0.37%) for heat rate. The maximum errors for the algorithm right after the change happened are 6.11% for power output and 2.8% for heat rate.

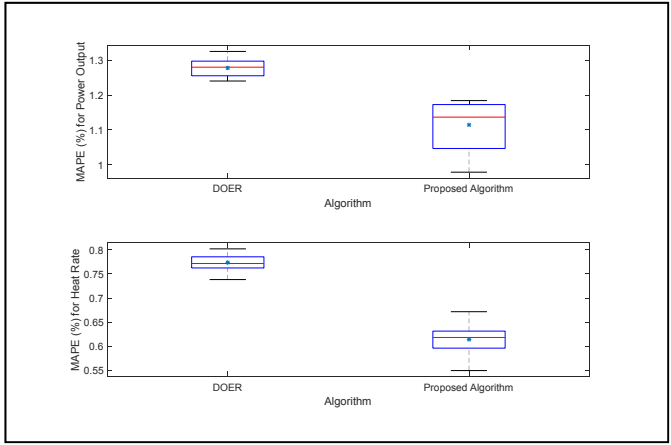

Fig. 6. Box-and-whisker plots that summarize the performance (MAPE) of DOER and the proposed algorithm on the real data set when maintenance action (water wash) is taken. The blue star ('*') represents the mean.

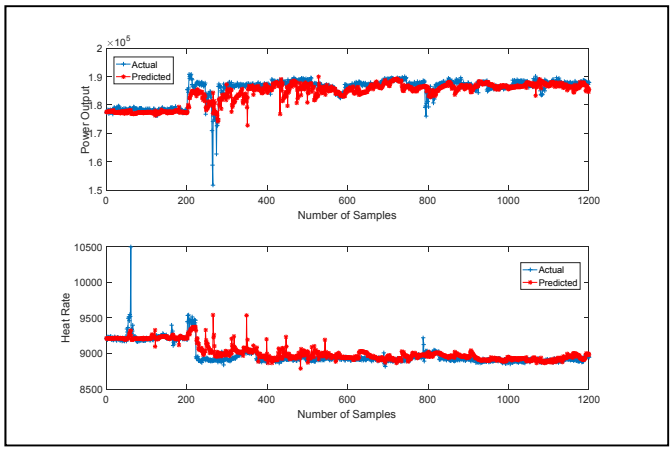

Fig. 7. Actual (blue line) and predicted (red line) power output and heat rate when water wash happened.

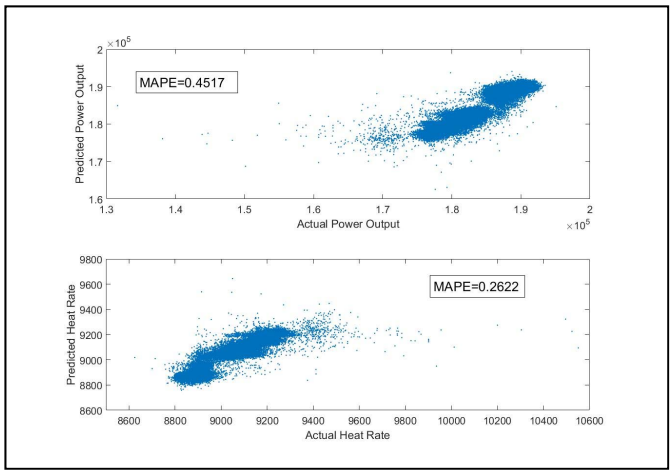

Fig. 8. Scatter plots of the predicted and actual power output (top) and heat rate (bottom) for the real data set.

Fig. 8 shows the scatter plots for both the predicted and actual power output and heat rate for the real data set. The MAPEs for power output and heat rate are 0.4517% and 0.2622%, respectively, which meet the <1% expectation. However, we also observe that there are some outliers with relatively large errors, which are caused by previously unseen data or noise or outliers in the data. A possible strategy to avoid such large errors in the forecasting is to allow some time lag for the updated model to be used for forecasting. In other words, we want to wait for a few samples to verify the performance of the currently updated model. This model is used to replace the current forecasting model only if its performance is better than that from the current forecasting model.

## IV. CONCLUSIONS

The paper presents an online ensemble-based approach for power plant performance modeling, which is important for plant real-time optimization and profit maximization. The continuous learning capability of the approach makes it possible to automatically update model in response to concept drifts due to component degradation, maintenance action, or operation change. The proposed approach consistently meets the requirements in real plant operation, with the overall MAPE prediction error < 1% on both simulated and real data. The approach also has the merit of scalability to different configured plants and easiness for implementation.